\documentclass[11pt]{article}
\usepackage[margin=1in]{geometry}
\usepackage{times}
\usepackage[T1]{fontenc}
\usepackage{amsmath,amssymb,amsfonts}
\usepackage{graphicx}
\usepackage{booktabs}
\usepackage{microtype}
\usepackage[numbers,sort&compress]{natbib}
\usepackage[colorlinks=true,linkcolor=blue,citecolor=blue,urlcolor=blue]{hyperref}
\usepackage{xspace}

\title{Small Enough to Know Everything:\\
The Fully-Enumerable Transformer as an Instrument\\
for the Science of Delayed Generalization}

\author{Yoshiyuki Ootani\\
Independent Researcher\\
\texttt{info@ootanl.com}}

\date{}

\begin{document}
\maketitle

\begin{abstract}
Tiny transformers trained on fully-enumerable tasks occupy an unusual position in
the study of grokking: every input can be evaluated, every generalization ceiling
can be computed exactly, and hundreds of seeds cost minutes. We argue that this
regime is best understood not as a scaled-down imitation of large-model training
but as a \emph{scientific instrument} with four capabilities that approximate
settings cannot offer: (a) theorem-grade generalization ceilings that turn
``can this split be generalized?'' into a computed, falsifiable bound;
(b) invariant-preserving task surgery, which manipulates a single structural
variable while provably fixing all others; (c) direct observation of every weight
without probes; and (d) survival-time statistics over 30--300 seeds that recast
``does not grok'' as a censored observation rather than a failure to replicate.
The obvious objection is that laws characterized at $10^4$ parameters may not mean
anything beyond them. We answer it with a preregistered conservation study: three
task-side laws established at 12K parameters---an exact recoverability-ceiling law,
a role-conflict delay law, and a weight-decay response law---are re-measured, under
an identical from-scratch protocol at 12K, 1M, and 50M parameters
($4{,}000\times$ span; 360 new runs plus a 44-run preregistered control arm). The ceiling law and the delay law are
conserved (0/144 Holm-corrected ceiling violations; Spearman $\rho \geq 0.75$
between role-conflict distance and grokking onset at every scale, permutation
$p < 10^{-4}$), while the weight-decay law deforms systematically: its response
steepens with scale in a direction consistent with larger models tolerating
stronger regularization.
Conservation was not assumed but tested against criteria frozen before data
collection, and the third relationship fails that test even at the original
scale, which is what tells us the test could have failed. The series scales the
model while the tasks stay enumerable, so what it licenses is specific: the
quantities this regime computes from a task keep predicting once the model has
outgrown the regime itself.
\end{abstract}

\section{Introduction}
\label{sec:intro}

Grokking---delayed generalization long after training accuracy
saturates~\citep{power2022grokking}---has become a standard lens on how neural
networks form structured solutions. Most quantitative work in this literature
already lives in a small, discrete world: modular arithmetic over ten digits,
permutation groups, sparse parities. What is rarely made explicit is that this
setting permits a mode of experimentation that is impossible anywhere else in deep
learning. When the input space has a few hundred cells and the model has $10^4$
parameters, \emph{nothing about the task or the model is hidden}: the entire
input--label table can be enumerated, the Bayes-optimal accuracy of any train/test
split can be computed exactly, and an experiment with three hundred seeds
finishes over lunch.

This paper makes two claims. The first is methodological: the fully-enumerable
regime should be treated as a scientific \emph{instrument}---a microscope for
delayed generalization---with four capabilities that we state precisely in
Section~\ref{sec:instrument} and that degrade or vanish the moment enumeration is
lost. Two of these capabilities (exact ceilings and task surgery) come from
enumerability of the task; the other two (full-weight observation and cheap
ensemble statistics) come from smallness of the model. The combination is what
makes the instrument: exact predictions on one side, exact measurements on the
other.

The second claim is empirical and answers the objection the first claim invites:
\emph{does anything measured at $10^4$ parameters constrain models three orders of
magnitude larger?} We formalize this as a conservation study
(Section~\ref{sec:design}). Three laws characterized at 12K
parameters~\citep{ootani2026grokking}---(L1) an exact recoverability-ceiling law,
(L2) a role-conflict delay law, and (L3) a non-monotone weight-decay response---are
re-measured at 12K, 1M, and 50M parameters under a single identical protocol:
same tasks, same splits, same tokenization, same optimizer, same budget, same
multi-seed survival analysis, all from scratch (the 12K arm is a
matched-protocol replication, not a reuse of the prior records). For each law we froze, before collecting any
cross-scale data, a three-way verdict rule: \emph{conserved}, \emph{deformed} (the
ordering survives but the quantitative form shifts systematically), or
\emph{broken}.

The results (Section~\ref{sec:results}) are: L1 conserved---the exact ceiling is
never exceeded in 144 Holm-corrected tests and the ceiling-ordering of grok rates
is monotone at every scale; L2 conserved---role-conflict distance predicts
grokking onset with Spearman $\rho \geq 0.75$ at every scale;
L3 deformed---the weight-decay response steepens with scale, and its descending
branch is absent throughout this from-scratch series, including at the original
12K scale, which demotes it from a law of the task family to a property of the
prior training setting. A 12K control arm run on GPU confirms that the
CPU$\to$GPU environment change does not move rates, isolating scale as the
manipulated variable.

We emphasize what conservation buys, and what it does not. It does not claim
that a 50M model is a big 12K model; mechanisms may differ, and our L2 scale
map (larger models are \emph{less} tolerant of role conflict, not more) shows
the mapping is not identity. Nor does it show that task-side quantities survive
the loss of task enumerability: the tasks here stay small enough to enumerate at
every scale, so the ceiling and $D_{\mathrm{role}}$ remain computable
throughout. What varies is the model, and with it the two capabilities that
come from smallness---full-weight observation and cheap ensembles---which are
gone by 50M in any practical sense. The claim is therefore precisely this:
\emph{quantities computed from the task by enumeration keep their predictive
force once the model has left the regime the instrument was built in}. That is
the license a microscope needs before its readings are quoted elsewhere;
extending it to tasks too large to enumerate is a separate question we take up
in Section~\ref{sec:limits}.

\section{The four capabilities of the fully-enumerable regime}
\label{sec:instrument}

Throughout, a task is a finite table $T\colon X \to Y$ with $|X| \sim 10^2$,
presented to a decoder-only transformer through a tokenization $E$, split into
train/test cells $S$, and trained with a fixed optimizer to a budget $B$.
``Fully enumerable'' means every element of $X$ is evaluated at every
checkpoint---there is no sampling in the measurement. The four capabilities, each
with a one-line empirical demonstration from prior
work~\citep{ootani2026grokking}, are:

\paragraph{(a) Theorem-grade ceilings (from enumerability).}
For a declared hypothesis class, the version space consistent with any training
split can be enumerated, so the expected accuracy of the Bayes-optimal predictor
on the held-out cells---the \emph{recoverability ceiling}, the best accuracy any
predictor could achieve given what the split reveals---is a computed number,
not an estimate. Ceilings turn ``the model failed to generalize'' into a
decidable claim: either the split was information-theoretically closed, or the
model underperformed an attainable bound. At 12K parameters, converged accuracy
tracks the computed ceiling with $r=0.95$ across graded task families, with zero
violations in 36 capacity-varied runs.

\paragraph{(b) Invariant-preserving task surgery (from enumerability).}
Because the task is a finite object in code, structural interventions can be
proven to change exactly one variable. Two examples: intercalated-swap
corruptions that provably preserve Latin-square structure while destroying an
additive-potential decomposition; and the \emph{rolex} construction used here,
which sets the two operand codebooks to permutations $\pi_a$ and
$\pi_b = \pi_a \circ \sigma$ for a random $(D{+}1)$-cycle $\sigma$, fixing both
marginal distributions while setting the role-conflict distance
$D_{\mathrm{role}} = d_{\mathrm{Cayley}}(\pi_a, \pi_b)$---how far the two
operands' codebooks are from agreeing---exactly. Under this surgery,
$D_{\mathrm{role}}$ predicts grokking onset with $\rho = 0.77$ ($n{=}60$).

\paragraph{(c) Full-weight observation (from smallness).}
A $10^4$-parameter model can be read in its entirety---every attention map on
every input, every neuron on the full input space---so hypotheses about internal
structure are checked by exhaustion rather than by probe proxies. This is the
regime in which complete mechanistic accounts of modular addition were first
obtained~\citep{nanda2023progress,zhong2023clock}; enumerability removes the
last sampling step from such analyses.

\paragraph{(d) Cheap ensemble statistics (from smallness).}
At minutes per run, grokking becomes a \emph{rate} science: Kaplan--Meier
survival curves over onset epochs, Wilson intervals over grok rates, and
right-censoring for runs that do not grok within budget---``does not grok'' is a
censored observation, not an anomaly. This matters because individual
trajectories are provably fragile: changing only the floating-point environment
flips the grok/no-grok outcome of $16$--$19\%$ of seeds while leaving ensemble
rates statistically unchanged. Claims below the rate level do not replicate; the
instrument is calibrated to the level at which they do.

None of these capabilities is new in isolation; Section~\ref{sec:related}
delineates what is. The point is their conjunction---and the fact that the
conjunction is exclusive to this regime. Approximate any one ingredient (sample
the input space, grow the model, run three seeds) and the corresponding
capability degrades from theorem to estimate.

\section{The conservation study}
\label{sec:design}

\subsection{Question and stakes}

The instrument is only interesting if its readings mean something outside it. We
therefore test: \emph{do task-side laws characterized at 12K parameters survive,
under an identical protocol, a $4{,}000\times$ scale increase?} A positive answer
licenses the enumerable regime as a model organism---playing for the laws of
delayed generalization the role that muP-style transfer plays for
hyperparameters~\citep{yang2021tensor}. A negative answer, under preregistered criteria,
would itself be a finding: the breaking point of task-side prediction.

\subsection{Protocol identity}

Everything except model size is held fixed across scales: the task tables, the
per-seed splits, the word-level digit tokenization, batch size 16, AdamW at
learning rate $3\times10^{-3}$, weight decay $0.1$ (except in L3, where weight
decay is the manipulated variable), a budget of $15{,}000$ epochs, and a
\emph{sustained} grokking criterion (test accuracy $\geq 0.70$ on two consecutive
full-space evaluations; the first such epoch is the onset). All runs are trained
from scratch, eliminating pretraining as a confound. Model configurations follow
a standard Llama-style stack at three sizes: 12K (hidden 16, 2 layers),
1M (hidden 128, 4 layers), and 50M (hidden 512, 12 layers). Each cell
receives $n{=}12$ seeds (L2: 3 task instances $\times$ 4 seeds). The cross-scale
series comprises 360 runs; per-seed JSON records, including full evaluation
trajectories, are released with the paper.

The prior study's 12K records were collected on CPU, while all arms of the
conservation series (including its 12K arm) run on GPU. A control compares 12K
GPU cells against the CPU records at matched budget. This matters because
floating-point environment changes are known to flip individual
seeds~\citep{ootani2026grokking}; the control confirms they do not move
\emph{rates} (Section~\ref{sec:results}).

\subsection{Three laws and frozen verdicts}

\paragraph{L1 (recoverability ceiling).} Task family \texttt{addpotc}$K$:
modular addition with $K \in \{0,1,2,3\}$ operand rows fully held out. Holding
out $K$ rows makes their $10 \times K$ test cells provably unrecoverable, giving exact
per-split ceilings $1.00, 0.71, 0.43, 0.14$ on the 30-cell test set.
\emph{Frozen predictions:} (i) no run exceeds its ceiling at any scale (the
ceiling is an information-theoretic quantity, hence model-independent);
(ii) grok rates order monotonically in the ceiling at every scale;
(iii) attainment---how closely runs approach the ceiling---declines with
capacity, extrapolating the decline observed across $6.5\times$ capacity at 12K.
\emph{Verdicts:} conserved $=$ (i)$\wedge$(ii); deformed $=$ (i) with a
systematic distortion of (ii); broken $=$ any Wilson-significant excess.

\paragraph{L2 (role-conflict delay).} Task family \texttt{rolex}$D$ with
$D \in \{0,4,8\}$, three instances each: exact Cayley distance between operand
codebooks, marginals fixed by construction. \emph{Frozen predictions:} the
ordering of onsets by $D_{\mathrm{role}}$ is conserved within each scale
(Spearman $\rho \geq 0.5$, censored runs ranked at budget); the absolute delay
may shift with scale, and estimating that shift---the scale map---is part of the
study's value. \emph{Verdicts:} conserved $=$ within-scale $\rho \geq 0.5$;
deformed $=$ ordering intact with systematic absolute shift; broken $=$ $\rho$
confidence interval crossing zero.

\paragraph{L3 (weight-decay response).} Task \texttt{add} at weight decay
$\in \{0.01, 0.1, 0.3\}$, a three-point coarse-graining of the inverted-U
response established at 12K, where intermediate weight decay maximizes the grok
rate. \emph{Frozen predictions:} the inverted-U shape is conserved; its peak may
shift. \emph{Verdicts:} conserved $=$ the middle point exceeds both ends with
Wilson separation; deformed $=$ the response becomes monotone over the tested
range while rates stay positive; broken $=$ zero rates everywhere or an inverted
ordering.

All criteria, cell lists, and predictions were frozen in a design document
committed (with hash) before any cross-scale run was launched.

\section{Results}
\label{sec:results}

\begin{figure}[t]
\centering
\includegraphics[width=\linewidth]{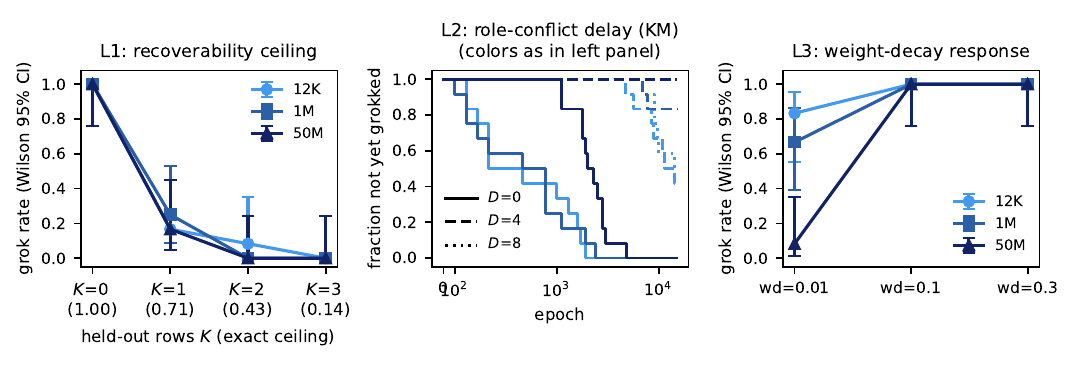}
\caption{The conservation map. \textbf{Left (L1):} grok rate versus number of
held-out rows $K$ (exact ceiling in parentheses) at 12K, 1M, and 50M parameters.
The ceiling-ordering is monotone at every scale, and rates beneath the ceiling
are statistically flat across scales. \textbf{Middle (L2):}
Kaplan--Meier curves of grokking onset for role-conflict distance $D \in
\{0,4,8\}$. $D{=}0$ groks at every scale; $D \geq 4$ is censored at the 15K-epoch
budget at 1M and 50M. \textbf{Right (L3):} grok rate versus weight decay. The
left arm of the weight-decay response sharpens with scale; the descending
arm has moved beyond $\mathrm{wd}{=}0.3$, a systematic deformation.}
\label{fig:map}
\end{figure}

\begin{figure}[t]
\centering
\includegraphics[width=0.5\linewidth]{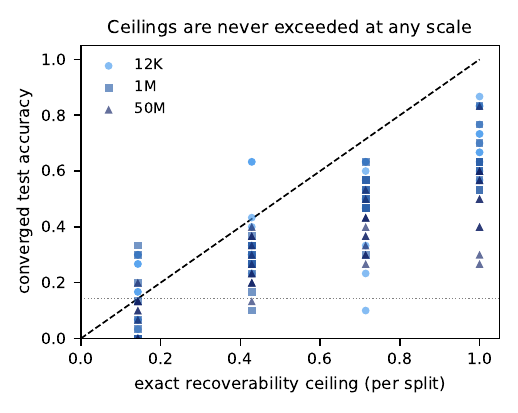}
\caption{Ceiling tracking across three orders of magnitude. Each point is one
run's converged (final) test accuracy against its split's exact recoverability
ceiling. No point lies significantly above the diagonal: 0 violations in 144
Holm-corrected exact binomial tests across 12K, 1M, and 50M (prior
capacity-sweep reference: 0/36).}
\label{fig:ceiling}
\end{figure}

\subsection{L1: conserved}

The exact ceiling is never exceeded. Testing every run's converged accuracy
against its split's ceiling with exact binomial tests (Holm-corrected across
all 144 runs of the from-scratch 12K/1M/50M series) yields zero violations
(Figure~\ref{fig:ceiling}), extending the earlier capacity-sweep record of
0/36. (The homogeneous binomial null is conservative here: the
true per-cell success probabilities are heterogeneous, and the matching
Poisson-binomial has smaller upper-tail mass, so the test can only under-report
excess---which strengthens a zero-violation finding.) The grok-rate ordering is
monotone in the ceiling at every scale (Figure~\ref{fig:map}, left): for
$K = 0 \to 3$, $12/12 \to 2/12 \to 1/12 \to 0/12$ at 12K,
$12/12 \to 3/12 \to 0/12 \to 0/12$ at 1M, and
$12/12 \to 2/12 \to 0/12 \to 0/12$ at 50M. Prediction (iii), by contrast, is
\emph{not} confirmed: at fixed $K{=}1$ the grok rates ($2/12$, $3/12$, $2/12$)
are statistically indistinguishable across scales, so the capacity-driven
attainment decline extrapolated from the 12K capacity sweep does not appear in
this series at $n{=}12$ resolution. We report the miss as the preregistration
requires; the conserved content of L1 is the exact bound and the ordering, and
capacity neither buys anything against the bound nor measurably costs
attainment beneath it here.

A methodological note worth one paragraph: if the excess test is run against the
\emph{best} accuracy over the trajectory rather than the converged accuracy, four
spurious ``violations'' appear, all in the $K{=}3$ cells where the test set is
noisiest. The best-over-trajectory statistic takes a maximum over $\sim$150
evaluations of a 30-cell test set, and its null distribution is not the single-look
binomial; the sustained criterion and converged-accuracy testing exist precisely
to remove this multiple-looks inflation. Small-test-set grokking claims based on
peak accuracy should be read with this failure mode in mind---a concern
formalized independently by the measurement-validity audit of
\citet{truong2026atgrok}.

\subsection{L2: conserved, with an informative scale map}

Within-scale Spearman correlation between $D_{\mathrm{role}}$ and onset
(censored runs at budget rank) is $\rho = 0.75$ at 12K, $\rho = 0.85$ at 1M,
and $\rho = 0.84$ at 50M ($n{=}36$ each; permutation $p < 10^{-4}$ at every
scale, $10^{5}$ label permutations---the asymptotic normal approximation is
invalid under this degree of censoring-induced tying, so we do not use it).
All three comfortably clear the frozen $\rho \geq 0.5$ criterion. (The earlier
12K measurement of $\rho = 0.77$ on a five-level $D$-grid is not directly
comparable to these three-level-grid values; the $\rho = 0.75$ above is the
matched-protocol replication.)

The scale map is the second finding (Figure~\ref{fig:map}, middle). At
$D{=}0$, the Kaplan--Meier median onset grows with scale (12K: 700 epochs;
1M: 900; 50M: 2{,}300). Away from $D{=}0$ the same direction appears as
collapsing rates: at $D{=}4$ the grok rate falls from $7/12$ at 12K (KM median
$14{,}200$, barely inside budget) to $2/12$ at 1M and $0/12$ at 50M; at
$D{=}8$, from $6/12$ at 12K to $0/12$ at both larger scales. Larger models are
less, not more, tolerant of role conflict per epoch of training. Whatever
intuition says scale should wash out a representational obstruction, the data
say the obstruction's cost grows.

\paragraph{Controls: neither learning rate nor budget explains the deficit.}
The natural objection is that the fixed protocol (learning rate
$3\times10^{-3}$, 15K epochs, chosen at 12K) is simply wrong for 50M, and the
$D{=}4$ failure is an optimization artifact rather than a scale effect. We
preregistered a control arm (design and verdict criteria frozen before the
runs): at 50M, \texttt{rolex4} was re-run at learning rates $10^{-3}$ and
$3\times10^{-4}$ (the direction muP-style scaling would favor for a wider
model; $n{=}12$ each, with $D{=}0$ sanity cells), and separately at a
$3\times$ budget of 45{,}000 epochs ($n{=}12$). At both alternative learning
rates the $D{=}0$ cells grok ($4/4$ and $4/4$) and training accuracy
saturates, yet $D{=}4$ remains at $1/12$ and $0/12$---below the 12K rate of
$7/12$ everywhere. At the tripled budget, $D{=}4$ is $0/12$ (best accuracy
$\leq 0.65$). Under the frozen criteria, the scale map stands: the 50M
role-conflict deficit survives learning-rate adjustment and a $3\times$
budget extension, and is not an optimization-mismatch artifact within the
tested range.

The decisive comparison here is the within-scale differential rather than the
learning-rate sweep on its own. Relaxation accounts of grokking predict a delay
that \emph{lengthens} as the learning rate falls (a clock scaling as
$1/(\eta\lambda)$ in the regime analyzed by~\citet{kim2026wdclock}), so failure
at a smaller $\eta$ is partly expected and cannot by itself exclude a different
optimal setting for 50M. What no fixed clock can produce is the gap we measure
at \emph{identical} optimizer, learning rate, weight decay, and budget: $D{=}0$
groks in all seeds run ($12/12$ at the base rate and $4/4$ in each sanity
cell) while $D{=}4$ groks in at most one of twelve, at every learning rate
tested. The obstruction is task-side, and it is the gap---not the absolute
delay---that the scale map tracks.

\subsection{L3: deformed}

The frozen conservation criterion fails, and fails informatively
(Figure~\ref{fig:map}, right). The left arm sharpens monotonically with scale:
at $\mathrm{wd}{=}0.01$ the grok rate falls $10/12 \to 8/12 \to 1/12$ from 12K
through 50M. But the descending right arm---reported at 12K under the prior
study's protocol~\citep{ootani2026grokking}---does not appear by
$\mathrm{wd}{=}0.3$ at \emph{any} scale in this from-scratch series ($12/12$ at
both $\mathrm{wd}{=}0.1$ and $0.3$ everywhere). Under the frozen rules the
verdict is \emph{deformed} at all three scales, including the matched-protocol
12K arm: the response stays non-negative and ordered, but the peak has widened
or shifted right. The two components of the deformation should be kept
apart: the right arm's absence appears already at 12K and is therefore
protocol-driven (from-scratch versus the prior study's setting), whereas the
left arm's sharpening is the genuinely scale-lawful signal. The data are consistent with larger models tolerating stronger weight decay for
delayed generalization within a fixed budget, though over a three-point grid
with both upper cells saturated at $12/12$ they cannot distinguish a
right-shifted peak from a true plateau---saturated cells at $n{=}12$ carry no
power for that comparison. Pinning the peak's location and scaling requires a
finer weight-decay grid, which we leave preregistered for the follow-up
(Section~\ref{sec:next}).

The left arm has a natural mechanistic reading in terms of the late-time
relaxation account of \citet{kim2026wdclock}, whose clock is inversely
proportional to $\eta\lambda$ in the weak-regularization regime: at
$\lambda{=}0.01$ relaxation is an order of magnitude slower than at
$\lambda{=}0.1$, so runs at the smallest weight decay are plausibly
right-censored by the 15K-epoch budget rather than incapable of generalizing.
Two caveats keep this interpretive rather than quantitative. The result is
derived for full-batch heavy-ball dynamics on linear (or locally quadratic)
models, whereas our runs use mini-batch AdamW on a transformer---a gap that
account itself flags as able to change the clock---and it describes the
late-time loss tail rather than the abrupt accuracy transition our sustained
criterion measures. Nor does it explain the missing right arm: a clock
monotone in $\lambda$ supplies no mechanism for one, but the prior study
observed a descending arm at 12K over this same grid, so that difference
remains protocol-driven. What the clock does supply is a model-side factor
orthogonal to the task-side quantities L1 and L2 measure; the two compose, and
separating them is part of what the instrument is for.

We state the consequence for L3 plainly rather than leaving it to be inferred.
A relationship that changes shape when pretraining is removed, before any
scaling is applied, is not on the same footing as L1 and L2: it is
protocol-sensitive, and the honest reading is that the inverted U was a
property of the prior setting rather than a law of the task family. We report
it under its frozen label because the label was fixed in advance, but the
substantive finding in L3 is the demotion---one of the three relationships we
carried into the study does not survive contact with a matched protocol, and
the two that do are the ones computed from the task rather than fitted to a
training regime.

\subsection{Control: the environment does not move rates}

Re-running 12K cells on GPU (matched budget) reproduces the CPU rates within
overlapping Wilson intervals (\texttt{addpotc0}: $12/12$ GPU vs.\ $3/3$ CPU;
\texttt{addpotc2}: $1/12$ vs.\ $0/3$). With $n{=}3$ on the CPU side this
control is corroborative rather than independently decisive; the primary
evidence that floating-point environments flip individual seeds but not rates
is the prior fragility study~\citep{ootani2026grokking}, which this arm is
consistent with. Together they isolate model scale as the manipulated variable
in the arms above.

\subsection{Verdict summary}

\begin{table}[h]
\centering
\small
\begin{tabular}{llll}
\toprule
Law & 12K finding & Cross-scale result & Verdict \\
\midrule
L1 ceiling & $r{=}0.95$; 0/36 excess & 0/144 excess; ordering monotone at all scales & \textbf{conserved} \\
L2 role conflict & $\rho{=}0.77$ (5-level grid) & $\rho{=}0.75/0.85/0.84$ (12K/1M/50M) & \textbf{conserved} \\
L3 weight decay & inverted-U & left arm amplified; peak widens right & \textbf{deformed} \\
\bottomrule
\end{tabular}
\caption{Preregistered verdicts. Two laws conserved, one systematically
deformed, none broken; the exact ceiling is violated nowhere at any scale.}
\label{tab:verdicts}
\end{table}

\section{What the instrument makes possible}
\label{sec:next}

Conservation converts the four capabilities from curiosities into leverage:
quantities the instrument computes from the task (exact ceilings,
$D_{\mathrm{role}}$, split identifiability) remain \emph{predictive} of models
far outside the size range in which the instrument's other readings are
available. The research program this enables is a
predictive one: a task-side function $\Phi(T, E, S, H)$ that outputs, before
training, the expected ceiling, the grok probability within budget, and a
predicted onset band. We commit to one concrete, falsifiable step rather than a
prospectus: a preregistration of $\Phi$---its functional form, coefficients
calibrated on the existing task atlas, and its predictions for five task
families never used in its construction, including one cross-domain family
(compositional binding in enumerable worlds~\citep{ootani2026microground})---will
be published with cryptographic hashes of the frozen predictions before the
confirmation runs are executed. The present paper's conservation map fixes the
scales at which those predictions are licensed to apply.

\section{Related work}
\label{sec:related}

\paragraph{What we do not claim as new.} Exhaustive evaluation of small
algorithmic tasks is the standard grokking setting from its
inception~\citep{power2022grokking,liu2022towards}. Complete
mechanistic analyses of small models are
established~\citep{nanda2023progress,zhong2023clock,chughtai2023toy}, grokking
appears well beyond algorithmic data~\citep{liu2022omnigrok}, and mechanistic
accounts of the delay itself exist at the level of training
dynamics~\citep{kumar2024grokking} and, most recently, as an exactly solvable
late-time relaxation whose clock scales as
$(1-\beta)/(\eta\lambda)$~\citep{kim2026wdclock}. The model-organism framing has a
distinguished precedent in muP hyperparameter transfer~\citep{yang2021tensor}.
Delayed-generalization curves have been traced in language models of
35M--130M parameters~\citep{muckatira2026pretraining}, establishing that the
\emph{phenomenon} exists at scale; our question is different---whether
\emph{quantitative task-side laws} transport---and to our knowledge no prior
work has tested law conservation under a frozen protocol across three orders of
magnitude. Closest to our L2 axis, \citet{howe2026priors,howe2026whatmakes}
shows that structure-specific representational priors \emph{causally} control
the grokking delay from the model side; our role-conflict law is the task-side
counterpart (a structural distance computed before training), and the two are
complementary rather than competing---a point the preregistered $\Phi$ program
(Section~\ref{sec:next}) is designed to make quantitative.

\paragraph{What is new here.} (1) The explicit statement of the enumerable
regime's four capabilities as an instrument, with the enumerability/smallness
attribution; (2) the treatment of generalization ceilings as computed,
falsifiable bounds tested by exact statistics rather than as informal
observations~\citep[cf.][]{varma2023explaining}; (3) systematic
invariant-preserving task surgery as a design pattern; (4) grokking as rate
science---survival curves with right-censoring over tens to hundreds of seeds,
which goes beyond the 1--5 seeds per configuration typical of the literature
and beyond generic multi-seed reporting; and (5) the conservation map
itself---the preregistered cross-scale test that licenses (1)--(4) beyond the
regime that generated them.

\section{Limitations}
\label{sec:limits}

The clearest boundary is the one this study does \emph{not} cross. We scaled
the model, not the task: at 12K, 1M, and 50M the same 100-cell tables are
enumerated exactly, so the ceiling and $D_{\mathrm{role}}$ were computable in
every arm. What the series establishes is that quantities obtained by
enumeration keep predicting once the model outgrows the regime where the
instrument's own model-side readings are available---not that they survive the
loss of enumerability itself. The complementary experiment is to hold the model
fixed and grow the task past the point where its input space can be exhausted,
replacing the exact ceiling with an estimated one and asking whether the
estimate retains its predictive force. Until that is run, the instrument's writ
runs exactly as far as enumeration does, and its role for natural language or
vision is upstream---supplying calibrated laws and preregistered predictors,
not direct measurement.

Three narrower limits. The series spans $4{,}000\times$ in parameters but uses
one architecture family and one optimizer, with one learning rate and one
budget in the main arms (the control arm varies both, for one task family) and
a fixed batch size throughout, which we did not manipulate. The 50M arm is a
single hardware environment, though the 12K control and prior fragility results
bound the plausible environment effect at the rate level. And conservation of
task-side laws does not imply conservation of mechanism; the L2 scale map
(delay grows with scale) is a reminder that the mapping across scales is lawful
but not trivial.

\section*{Acknowledgments and AI-use disclosure}

Claude (Anthropic) was used, under the author's direction, in the preparation
of this work: drafting and revising the manuscript text, implementing the
experiment, analysis, and figure code, executing the preregistered runs, and
checking literature and internal consistency. The research questions, study
design, frozen prediction criteria, and interpretation of results are the
author's, and the author reviewed, verified, and takes full responsibility
for all content, including every quantitative claim.

\bibliographystyle{unsrtnat}
\bibliography{refs}

\end{document}